\pdfoutput=1

\documentclass[11pt]{article}

\usepackage[final]{acl}

\usepackage{times}
\usepackage{latexsym}

\usepackage[T1]{fontenc}

\usepackage[utf8]{inputenc}

\usepackage{microtype}

\usepackage{inconsolata}

\usepackage{graphicx}

\usepackage[textsize=small]{todonotes}
\usepackage{xcolor}
\usepackage{soul}

\usepackage{amsmath}
\usepackage{booktabs}
\usepackage{multirow}
\usepackage{arydshln}
\usepackage{xcolor}
\usepackage{xspace}
\usepackage{algorithm}
\usepackage{algorithmic}
\newcommand{\proj}{Mamba-Shedder\xspace}
\usepackage{hyperref}

\usepackage{colortbl}
\definecolor{customcolor1}{HTML}{fefcef}
\definecolor{customcolor2}{HTML}{f4faf4}

\title{
\proj: Post-Transformer Compression for Efficient Selective Structured State Space Models}

\author{
 \textbf{J. Pablo Muñoz \textsuperscript{1}}\thanks{
 Co-first authors. 
 },
 \textbf{Jinjie Yuan\textsuperscript{2}}\footnotemark[1],
 \textbf{Nilesh Jain\textsuperscript{1}}
\\
 \textsuperscript{1}Intel Labs,
 \textsuperscript{2}Intel Corporation\\
 \small{
 \{pablo.munoz, jinjie.yuan, nilesh.jain\}@intel.com
 }
}

\begin{document}
\maketitle

\begin{abstract}

Large pre-trained models have achieved outstanding results in sequence modeling. The Transformer block and its attention mechanism have been the main drivers of the success of these models. Recently, alternative architectures, such as Selective Structured State Space Models (SSMs), have been proposed to address the inefficiencies of Transformers. This paper explores the compression of SSM-based models, particularly Mamba and its hybrids. We study the sensitivity of these models to the removal of selected components at different granularities to reduce the model size and computational overhead, thus improving their efficiency while maintaining accuracy. The proposed solutions, collectively referred to as \textbf{\proj}, achieve a speedup of up to 1.4x during inference, demonstrating that model efficiency can be improved by eliminating several redundancies with minimal impact on the overall model performance. The code is available at \href{https://github.com/IntelLabs/Hardware-Aware-Automated-Machine-Learning}{https://github.com/IntelLabs/Hardware-Aware-Automated-Machine-Learning}.
\end{abstract}

\section{Introduction}

We have seen an outstanding increase in the number of Transformer-based models \cite{Transformer_NIPS2017} developed to tackle tasks from Natural Language Processing (NLP) and other domains \cite{46840_image_transformer, dosovitskiy2021imageworth16x16words, vivit_9710415, gong21b_interspeech} due to their effectiveness at modeling sequences. However, these models also present critical efficiency challenges. For example, the cost of training these models scales quadratically in the sequence length. In the generation stage, Transformers, in their original form, require large caches to store the previously seen tokens. Several variants of Transformers have been proposed to address these efficiency challenges, but researchers have also explored alternative post-Transformer architectures to address these limitations. \emph{Structured state space models (SSMs)}, e.g., S4 \cite{gu2022efficientlyS4}, followed by \emph{Selective state space models}, e.g., Mamba \cite{mamba1, mamba2} have been proposed as efficient alternatives that achieve training time with linear scaling in sequence length, and during generation, maintain constant state size. 

Model compression methods, e.g., pruning and quantization, have been broadly explored and applied to Transformer-based models. However, more must be done to explore compression in their structured state space counterparts. This paper explores the pruning of these alternative architectures, presenting results that provide insights into potential opportunities to increase their efficiency without sacrificing accuracy. The rest of the paper discusses the following contributions:  
\begin{itemize}
    \item A pruning solution, \textbf{\proj}, which targets structures in selective structured state space models, improving their computational and memory efficiency.
    \vspace{-0.2cm}
    \item Comprehensive experiments to determine the tolerance of SSM-based models to the removal of their structures.
    \vspace{-0.2cm}
    \item Insights on how the differences in the SSM building blocks and their interaction with Transformer blocks in hybrid models affect the trade-off between efficiency and accuracy.        
\end{itemize}
\vspace{-0.1cm}
The following content is organized as follows: Section \ref{sec:preliminaries} provides the reader with details of the alternative architectures utilized in our study and popular strategies for element removal in large models. Section \ref{sec:methodology} describes methods to study network pruning in Mamba and hybrid architectures. Section \ref{sec:experiments} presents the results of our experiments and ablation studies, and we offer concluding remarks in Section \ref{sec:conclusion}. A Related Work section is included in the Appendix.

\section{Preliminaries}
\label{sec:preliminaries}

\subsection{State Space Models}
State space models (SSMs) have a long history of modeling sequences and dynamic systems. Recently, \emph{structured} SSMs, e.g., S4 \cite{gu2022efficientlyS4}, have been proposed as an alternative to Transformers because of their efficient capabilities for mapping input to output signals. When dealing with discrete sequences as in Natural Language Processing (NLP), the parameters $\boldsymbol{A}$, $\boldsymbol{B}$ and $\boldsymbol{C}$ of these models are discretized to transform an input sequence, $x_t$, and hidden state, $h_t$, to obtain the output sequence, $y_t$.  
It can be formalized as:
\begin{equation}
h_t = \boldsymbol{A} h_{t-1} + \boldsymbol{B} x_t, y_t = \boldsymbol{C}^\top h_t. 
\label{eq:ssm}
\end{equation} 

\label{sec:mamba_prelim}
\paragraph{Mamba: Selective State Space Models} 
S4 and other structured SSMs are linear time-invariant (LTI), i.e., their parameters are fixed, limiting their effectiveness for sequence modeling. For instance, structured state space models fail in many content- and context-based reasoning tasks. These limitations have motivated the development of time-varying alternatives, e.g., Mamba \cite{mamba1}, which incorporate selection mechanisms and are suitable for solving tasks previously SSM generations failed. Specifically, Mamba's SSM module, S6, allows its parameters to depend on the input, thereby modifying the formulation from time-invariant to time-varying. A second improvement proposed in Mamba compared to previous SSMs is a hardware-aware algorithm that speeds up execution while reducing memory IOs.

\begin{figure*}
  \centering
  \includegraphics[width=\linewidth]{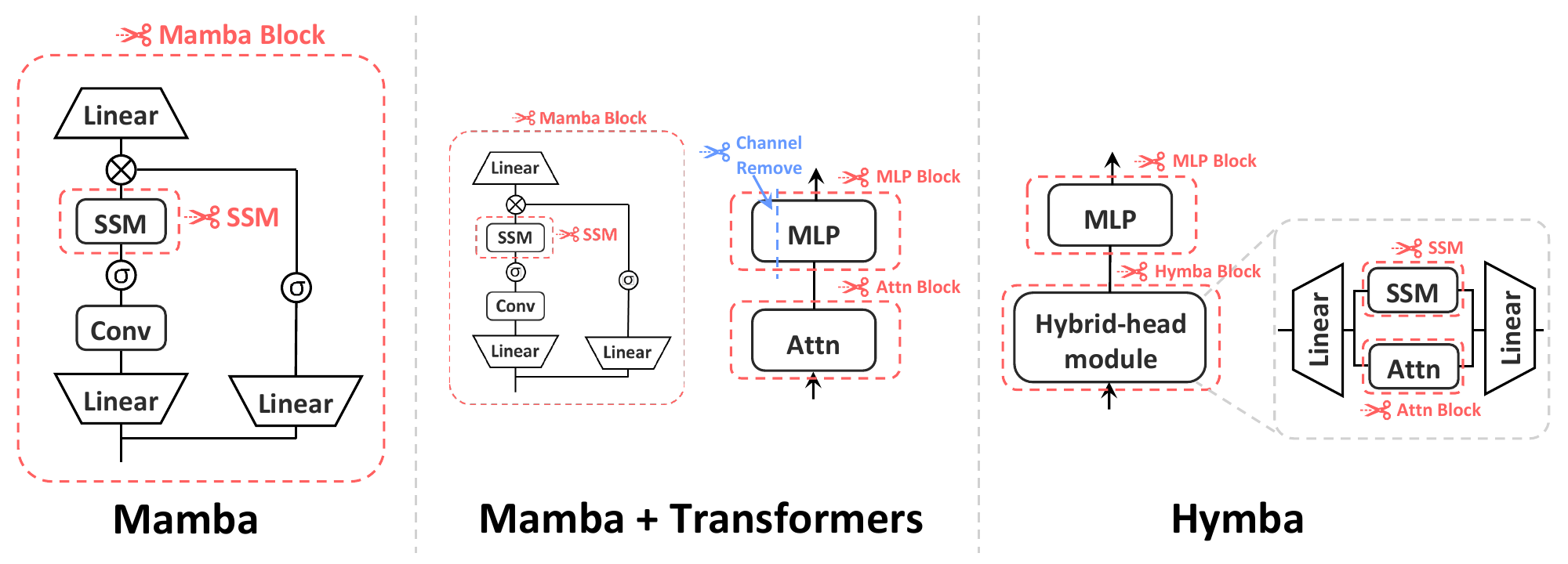}
  \caption{
  Overview of \proj. This figure illustrates the pruning strategy for three types of Mamba-based models. The first type includes Mamba models such as Mamba-1 \cite{mamba1}, Mamba-2 \cite{mamba2}, and Falcon-Mamba \cite{zuo2024falcon}. The second type comprises Mamba + Transformers architectures, including Zamba \cite{glorioso2024zambacompact7bssm}. The third type is Hymba \cite{dong2024hymba}, a novel architecture with hybrid heads. Red dashed lines indicate potential removal. In Transformers, channel pruning can also be applied to MLP block (width pruning). 
  }
\label{fig:mamba_shedder}
\end{figure*}

Furthermore, Mamba-2 \cite{mamba2} improves the original Mamba architecture by proposing \emph{state space duality (SSD)}, which improves its efficiency on hardware accelerators compared to S6. This improvement is achieved by changing the \emph{state matrix}, $\boldsymbol{A}$, which directly controls the latent state, $h$. $\boldsymbol{A}$ is modified from being structured as a diagonal matrix to a formulation that utilizes a scalar-times-identity structure.  

Additionally, Mamba-2 introduces the concept of heads in SSMs inspired by how multi-head attention (MHA) works and implementing a grouped-value attention (GVA) head structure.  
Overall, the Mamba-2 architecture, with its SSD core component, allows for improved parallelism of the block's projections. 

\paragraph{Mamba block} Mamba models comprise several blocks stacked after each other. Figure \ref{fig:mamba_shedder} on the left illustrates a single Mamba block. Each block has the selective SSM mechanism (S6 for Mamba-1 and SSD for Mamba-2) at its core, placed within a larger structure that combines a gated multilayer perceptron (MLP), a convolution, and SILU activation functions \cite{ELFWING20183_silu}.   

For more details about selective structured state space models, we refer the reader to \citet{mamba1} and \citet{mamba2}.

\subsection{Hybrid Models} 
Lately, new models have been proposed that achieve the best of both worlds (Transformers and Selective SSMs) by proposing architectures with both classes of blocks. Zamba
\cite{glorioso2024zambacompact7bssm} is one example of such a hybrid model. It combines the strengths of Mamba's backbone and the efficiency of selective SSMs with a shared Transformer block that incorporates Transformers' powerful in-context learning capabilities. The \emph{shared attention} mechanism, in which two attention blocks are reused and interleaved in an ABAB pattern throughout the network, is a characteristic innovation of Zamba. This model also applies LoRA adapters \cite{hu2022lora} to the shared MLP blocks, achieving specialization when interacting with the affected layers, memory efficiency, and faster inference with reduced computational overhead. 

Another example of a hybrid model is Hymba \cite{dong2024hymba}. This model takes a different approach than Zamba, proposing an entirely new hybrid-head module, illustrated in Figure \ref{fig:mamba_shedder} on the right, in which the SSM and Attention mechanisms contribute in parallel to the sequence modeling. Additionally, Hymba benefits from group query attention, cross-layer KV cache sharing, and learnable meta-tokens, resulting in higher throughput, reduced memory requirements, and competitive performance compared to models of similar size. 

\subsection{Model Pruning}

A popular model compression technique, \emph{pruning} \cite{LeCunBrainDamage}, has been effectively used to reduce the size of deep learning models and improve their efficiency. Network pruning operates at two levels: (1) \emph{Unstructured pruning} identifies the importance of individual weights that can be masked to minimize their impact on overall model behavior.  
At a different level, (2) \emph{structured pruning} focuses on removing more significant structural components of the model, such as whole Transformer blocks \cite{men2024shortgptlayerslargelanguage}, or reducing the granularity to target subcomponents of these layers \cite{zhong2024blockprunerfinegrainedpruninglarge, muñoz2025multiprunerbalancedstructureremoval}. Other dimensions for pruning include groups of channels in the Transformer's MLPs or heads from the MHA layer.  In this paper, the focus is solely on structured pruning applied to Mamba-based models.

Next, we discuss \proj's methodology to study redundancies in Mamba and hybrid models. 

\section{Methodology}
\label{sec:methodology}

\begin{figure*}[ht]
    \centering
    \begin{minipage}{0.30\textwidth}
        \centering
        \includegraphics[width=\textwidth]{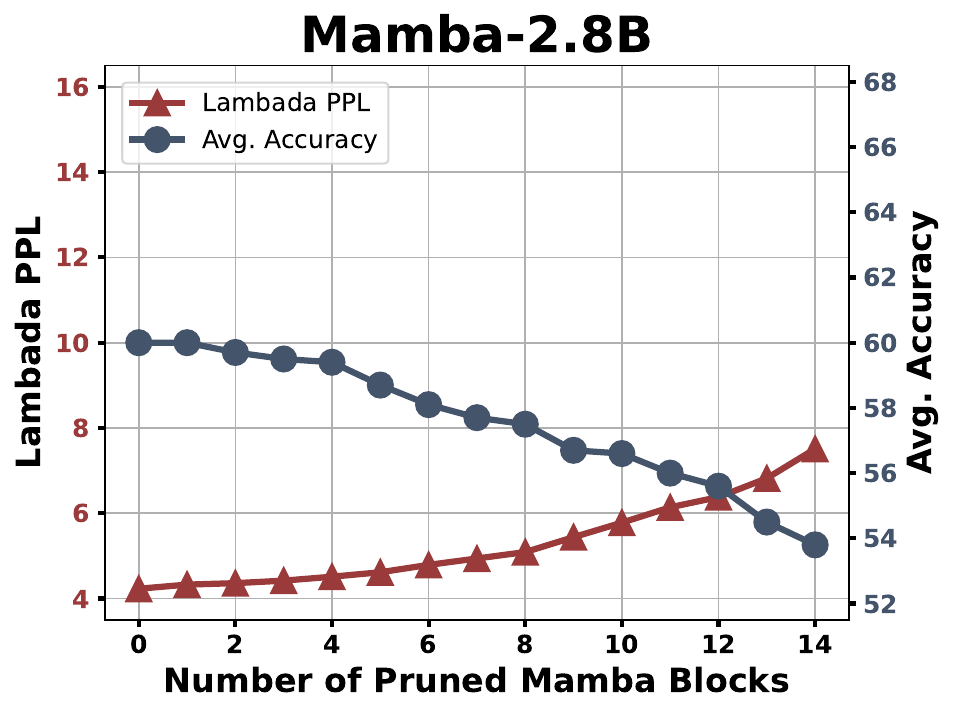}
    \end{minipage}
    \hfill
    \begin{minipage}{0.30\textwidth}
        \centering
        \includegraphics[width=\textwidth]{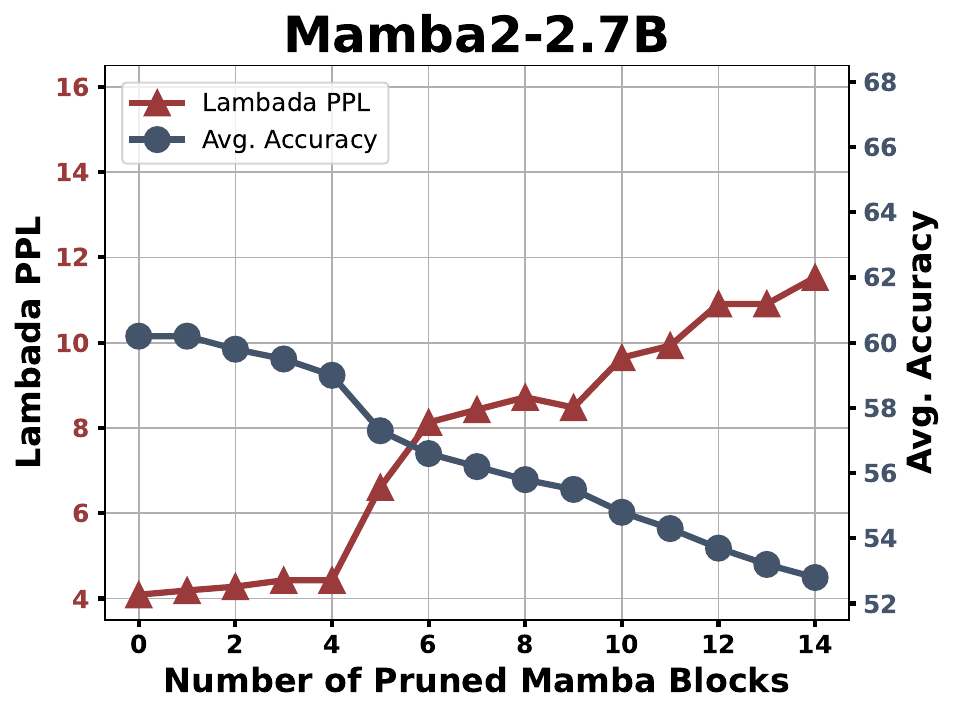}
    \end{minipage}
    \hfill
    \begin{minipage}{0.30\textwidth}
        \centering
        \includegraphics[width=\textwidth]{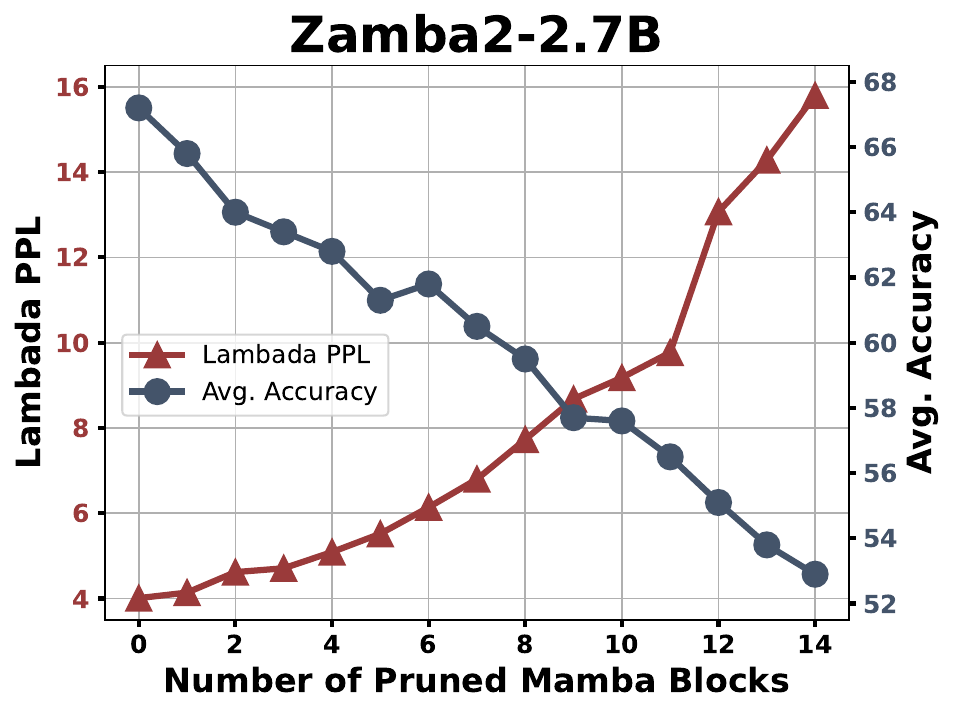}
    \end{minipage}
    \caption{Pruning Mamba blocks. \emph{Avg. Accuracy} indicates the average accuracy for seven tasks. The model composed of Mamba 1 blocks (left) can tolerate the removal of entire blocks without significantly increasing its perplexity or decreasing accuracy compared to Mamba-2 and Zamba-2. In all three models, removing each Mamba block reduces 0.04B parameters from the model. 
    These are \emph{training-free} results, and drops in accuracy can be reduced by a subsequent fine-tuning stage (\S 4.5). 
    }
    \label{fig:mamba_block_pruning}
\end{figure*}

\begin{table*}[!t]
    \setlength{\tabcolsep}{2.5pt}
    \centering
    \scriptsize
    \renewcommand\arraystretch{1.2}
    \begin{tabular}{llccccccccccc}
    \toprule
        \multirow{2}{*}{\textbf{Model}} & \multirow{2}{*}{\textbf{Method}} & \textbf{Num. of Pruned} & \multirow{2}{*}{\textbf{Ratio}} & \textbf{Lambada} & \multirow{2}{*}{\textbf{Lambada}} & \multirow{2}{*}{\textbf{HellaS}} & \multirow{2}{*}{\textbf{PIQA}}  & \multirow{2}{*}{\textbf{ARC-e}} & \multirow{2}{*}{\textbf{ARC-c}} & \multirow{2}{*}{\textbf{WinoG}}  & \multirow{2}{*}{\textbf{OBQA}} & \multirow{2}{*}{\textbf{Average}} \\ 
         &  & \textbf{\textcolor{red}{Mamba Blocks}} &&  \textbf{PPL ($\downarrow$)} \\ 
    \midrule
        \multirow{3}{*}{\textbf{Mamba-2.8B}} & Dense & 0 / 64 & 0\% &  4.23&	69.2	&66.1	&75.2	&69.7	&36.3&	63.5	&39.6	&59.9  \\ 
        \cdashline{2-13}
        & \multirow{2}{*}{Mamba Block Pruning} & \cellcolor{customcolor1} 7 / 64 & \cellcolor{customcolor1}10.43\%&	\cellcolor{customcolor1}\underline{\textbf{4.94$_\text{+0.71}$}}	&\cellcolor{customcolor1}65.8&	\cellcolor{customcolor1}63.7	&\cellcolor{customcolor1}73.8	&\cellcolor{customcolor1}68.0	&\cellcolor{customcolor1}33.5&	\cellcolor{customcolor1}62.5&	\cellcolor{customcolor1}36.8&	\cellcolor{customcolor1}\underline{\textbf{57.7$_\text{-2.2}$}}  \\ 
        &  & \cellcolor{customcolor1}14 / 64 &  \cellcolor{customcolor1}20.86\%	& \cellcolor{customcolor1}\underline{\textbf{7.51$_\text{+3.28}$}} &	\cellcolor{customcolor1}58.9&	\cellcolor{customcolor1}57.6&	\cellcolor{customcolor1}71.0	&\cellcolor{customcolor1}62.7	&\cellcolor{customcolor1}32.0	&\cellcolor{customcolor1}61.1&	\cellcolor{customcolor1}33.2&	\cellcolor{customcolor1}\underline{\textbf{53.8$_\text{-6.1}$}} \\ 
    \midrule
        \multirow{3}{*}{\textbf{Mamba2-2.7B}} & Dense & 0 / 64 & 0\% &  4.10&	69.7&	66.6	&76.4&	69.6&	36.4	&64.0	&38.8	&60.2  \\ 
        \cdashline{2-13}
        & \multirow{2}{*}{Mamba Block Pruning} & 7 / 64 & 10.42\%	&\textbf{8.43$_\text{+4.33}$}	&53.0&	63.8&	73.9	&66.6&	36.4&	64.5&35.0	&\textbf{56.2$_\text{-4.0}$}  \\ 
        &  & 14 / 64 &  20.83\%	&\textbf{11.53$_\text{+7.43}$}&	47.0	&59.4	&71.1	&60.6	&35.6	&60.8	&35.0&	\textbf{52.8$_\text{-7.4}$} \\ 
    \midrule
        \multirow{3}{*}{\textbf{Zamba2-2.7B}} & Dense & 0 / 54 & 0\% & 4.01&	69.7	&77.0	&79.8	&77.5&	48.5	&72.1&	45.8&	67.2  \\ 
        \cdashline{2-13}
        & \multirow{2}{*}{Mamba Block Pruning} & 7 / 54 & 10.38\%  &	\textbf{6.80$_\text{+2.79}$}	 & 58.9	 & 69.7 &	77.0	 &69.8	 &39.6 &	67.0 &	41.8	 &\textbf{60.5$_\text{-6.7}$}  \\ 
        &  & 14 / 54 & 20.77\%&	\textbf{15.8$_\text{+11.79}$}	&44.3&	62.8	&72.7	&54.3	&34.5&	64.3&	37.2	&\textbf{52.9$_\text{-14.3}$}  \\ 
    \bottomrule
    \end{tabular}
\caption{
Detailed results of \proj with \emph{training-free} Mamba block pruning. Lambada, HellaS, PIQA, ARC-e, ARC-c, WinoG, and OBQA represent their respective accuracies. \underline{Underlined} numbers indicate the smallest average accuracy gap with the dense model under the same level of pruning. 
}   
\label{tab:mamba_block_pruning}
\end{table*}

Due to the large sizes of current state-of-the-art sequence models, \proj requires an efficient strategy to identify structures that can be removed without significantly affecting the model's accuracy. We approach this problem using a training-free approach, in which the least essential elements are considered for removal. Similar strategies have been explored in Transformer-based large language models \cite{ashkboos2024slicegpt, men2024shortgptlayerslargelanguage, zhong2024blockprunerfinegrainedpruninglarge}. However, to our knowledge, no study explores the removal of structures in Selective Structured State Space models. \proj conducts structure removal of Mamba models and their hybrid variants at different granularities. As illustrated in the left of Figure \ref{fig:mamba_shedder}, in the case of models with only Mamba blocks, we explore the iterative removal of entire Mamba blocks (\S \ref{sec:mamba_prelim}), or their SSM subcomponents, either S6 or SSD modules depending on the version of Mamba (Figure \ref{fig:mamba_shedder}).

The proponents of the Mamba architecture do not provide a rationale for the number of Mamba blocks required to build robust models, opening an opportunity for \proj to investigate whether some components might be redundant and hence removed from the model with a minor impact in accuracy. 

In addition to these components, in the case of hybrid models that also contain Transformer blocks (middle of Figure \ref{fig:mamba_shedder}), we also explore the removal of entire Transformer blocks or their subblocks: multilayer perceptrons (MLP) modules and multihead attention (MHA) modules.
In hybrid models, \proj also explores the removal of structures at a finer granularity by targeting groups of channels in the MLP's linear layers, i.e., based on a channel group size, $g$, \proj explores the removal of $n g$ channels, where $n$ is the number of groups that could be removed based on their impact of the overall model performance. 

\begin{algorithm}[H]
  \caption{Block / Module Pruning}
  \small   
  \label{alg:block_pruning}
  \textbf{Input:} Set of blocks/modules $\mathcal{M}$ from a model $m$, Calibration dataset $\mathcal{C}$, Metric $\phi$, Target pruning steps $t$.
  \\
    \textbf{Output:} Pruned model $m^*$
  \begin{algorithmic}[1]
    \FOR{$k \leftarrow 1$ \TO $t$}
        \FORALL{$M_i \in \mathcal{M}$}
            \STATE $S_{i} \gets \text{Importance}(M_i, m, \mathcal{C}, \phi)$
        \ENDFOR
        \STATE $M_{\text{min}} \gets \arg\min_{M_i \in \mathcal{M}} S_{i}$
        \STATE $\mathcal{M} \gets \mathcal{M} \setminus \{M_{min}\}$ \hfill \(\triangleright\) \textcolor{red}{Block/Module Pruning}
    \ENDFOR
    \RETURN $m^* \text{ with the remaining blocks/modules in } \mathcal{M}$
  \end{algorithmic}
\end{algorithm}

Algorithm \ref{alg:block_pruning} details the procedure to remove entire structures, e.g., Mamba or Transformer blocks, MLPs, MHA, or SSM modules. Given a set $\mathcal{M}$ of structures selected for potential removal, a proxy data set $C$ and a metric $\phi$ are used to measure the importance of an individual structure and the impact of removing it from the model \cite{zhong2024blockprunerfinegrainedpruninglarge}. In addition to entire structures, \proj follows the same logic to remove channel groups as detailed in Algorithm \ref{alg:channel_pruning}.

\begin{algorithm}[H]
  \caption{MLP Channel Pruning
  }
  \small   
  \label{alg:channel_pruning}
  \textbf{Input:} Set of MLP blocks $\mathcal{M_{\text{MLP}}}$ from a model $m$, Calibration dataset $\mathcal{C}$, Metric $\phi$, Target pruning steps $t$, MLP channel group size $g$.
  \\
    \textbf{Output:} Pruned model $m^*$
  \begin{algorithmic}[1]
    \FOR{$k \leftarrow 1$ \TO $t$}
        \FORALL{$M_i \in \mathcal{M}_{\text{MLP}}$}
            \STATE $S_{i} \gets \text{Importance}(M_i[:, \text{:-}g], m, \mathcal{C}, \phi)$
        \ENDFOR 
        \STATE $M_{\text{min}} = \arg\min_{M_i \in \mathcal{M}_{\text{MLP}}} S_i$
        \STATE $M_{\text{min}} = M_{\text{min}}[:, \text{:-}g]$ \hfill \(\triangleright\) \textcolor{red}{Channel Pruning}
    \ENDFOR
    \RETURN $m^* \text{ with the altered MLP blocks in } \mathcal{M}$
  \end{algorithmic}
\end{algorithm}

Depending on the pruning objective, \proj might treat these pruning targets in isolation, but Section \ref{sec:experiments} also presents the results of configurations in which \proj sequentially prunes larger structures (e.g., Mamba blocks) and, at a later stage, smaller components, e.g., SSM modules in the remaining Mamba blocks. Future work will explore larger search spaces with more complex configurations of candidate structures for removal. For example, the importance of Mamba blocks and their SSM modules can be assessed in the same pruning iteration.  

\section{Experiments}
\label{sec:experiments} 

We evaluate \proj and study the removal of structures from SSM-based models utilizing several open-source models and datasets. We analyze their absolute and relative drop in accuracy and quantify the inference speedup obtained by the pruned models. Next, we discuss the resources utilized for our experiments and details of our setup and results. 

\begin{figure*}[ht]
    \centering
    \begin{minipage}{0.30\textwidth}
        \centering
        \includegraphics[width=\textwidth]{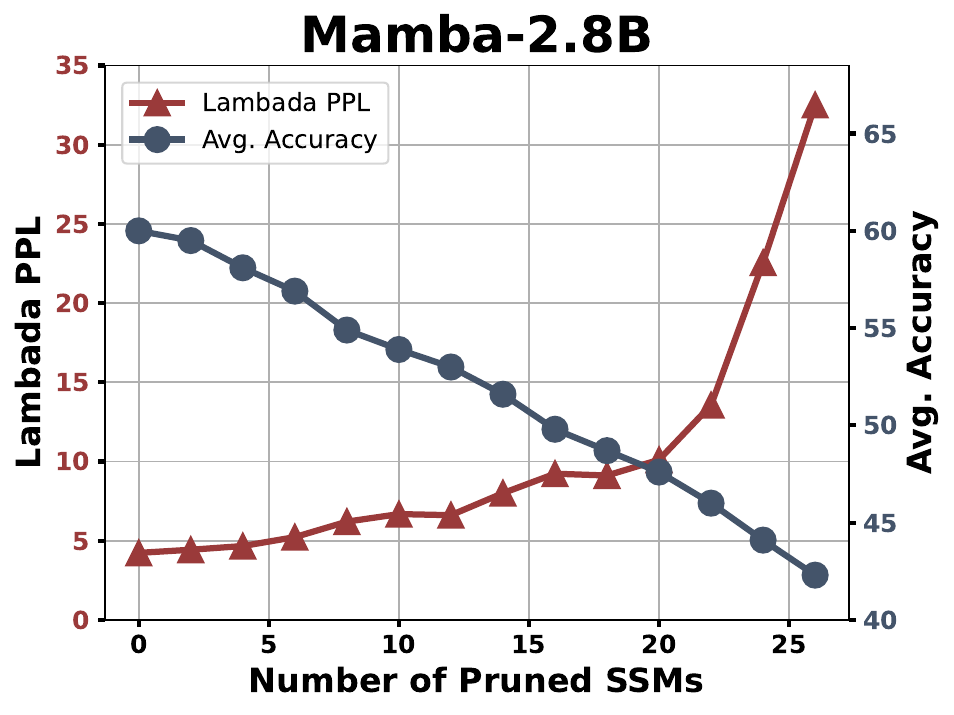}
    \end{minipage}
    \hfill
    \begin{minipage}{0.30\textwidth}
        \centering
        \includegraphics[width=\textwidth]{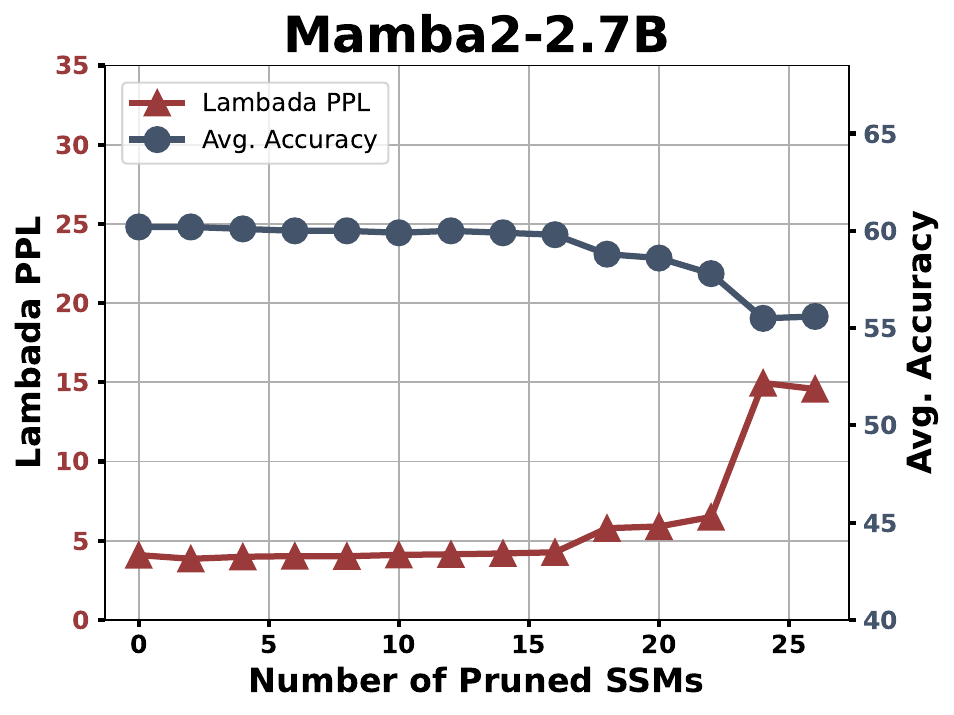}
    \end{minipage}
    \hfill
    \begin{minipage}{0.30\textwidth}
        \centering
        \includegraphics[width=\textwidth]{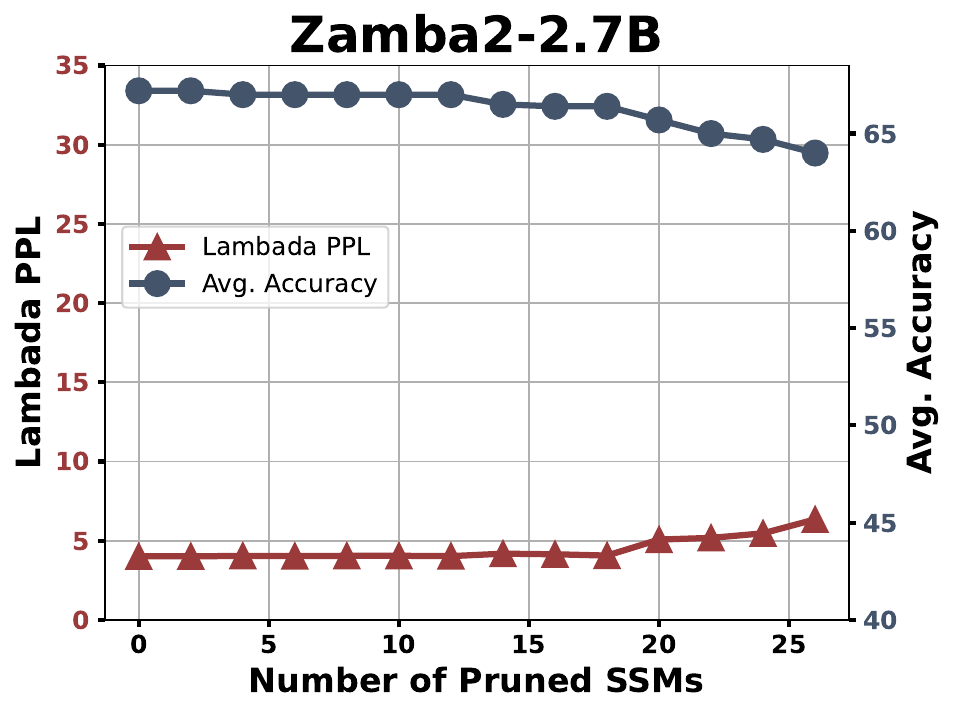}
    \end{minipage}
    \caption{Pruning SSM (S6 and SSD modules). Mamba-2.8B and Mamba2-2.7B have 64 SSM modules, while Zamba2-2.7B has 54 SSM (SSD) modules. \emph{Avg. Accuracy} is for the seven tasks evaluated.
    }
    \label{fig:ssm_pruning}
\end{figure*}

\begin{table*}[!t]
    \setlength{\tabcolsep}{4.5pt}
    \centering
    \scriptsize
    \renewcommand\arraystretch{1.2}
    \begin{tabular}{llcccccccccc}
    \toprule
        \multirow{2}{*}{\textbf{Model}} & \multirow{2}{*}{\textbf{Method}} & \textbf{Num. of} & \textbf{Lambada} & \multirow{2}{*}{\textbf{Lambada}} & \multirow{2}{*}{\textbf{HellaS}} & \multirow{2}{*}{\textbf{PIQA}}  & \multirow{2}{*}{\textbf{ARC-e}} & \multirow{2}{*}{\textbf{ARC-c}} & \multirow{2}{*}{\textbf{WinoG}}  & \multirow{2}{*}{\textbf{OBQA}} & \multirow{2}{*}{\textbf{Average}} \\ 
         &  & \textbf{Pruned \textcolor{red}{SSMs}} & \textbf{PPL ($\downarrow$)} \\ 
    \midrule
        \multirow{4}{*}{\textbf{Mamba-2.8B}} & Dense & 0 / 64  &  4.23&	69.2	&66.1	&75.2	&69.7	&36.3&	63.5	&39.6	&59.9  \\ 
        \cdashline{2-12}
        & \multirow{3}{*}{SSM Pruning} & 16 / 64 &\textbf{9.23$_\text{+5.00}$} &	55.2	&52.1&	68.1&	57.8	&28.4&	55.6&	31.6&	\textbf{49.8$_\text{-10.1}$}	\\ 
        &  & 20 / 64 &	\textbf{10.10$_\text{+5.87}$}	&57.1	&48.2&	65.5	&50.9	&25.9&	56.0&	29.4&	\textbf{47.6$_\text{-12.3}$} \\ 
        &  & 24 / 64 	&\textbf{22.55$_\text{+18.32}$}&	44.4&	43.2	&64.4&	47.4&	25.8&	53.6&	29.8&	\textbf{44.1$_\text{-15.8}$} \\ 
    \midrule
        \multirow{4}{*}{\textbf{Mamba2-2.7B}} & Dense & 0 / 64  &  4.10&	69.7&	66.6	&76.4&	69.6&	36.4	&64.0	&38.8	&60.2  \\ 
        \cdashline{2-12}
        &   \multirow{3}{*}{SSM Pruning} & \cellcolor{customcolor1} 16 / 64 	& \cellcolor{customcolor1} \textbf{4.26$_\text{+0.16}$}	&\cellcolor{customcolor1} 66.9	&\cellcolor{customcolor1} 66.1	&\cellcolor{customcolor1} 76.4&	\cellcolor{customcolor1} 68.6	&\cellcolor{customcolor1} 37.2&	\cellcolor{customcolor1} 64.0&	\cellcolor{customcolor1} 39.2	& \cellcolor{customcolor1} \underline{\textbf{59.8$_\text{-0.4}$}} \\ 
        &  & \cellcolor{customcolor1} 20 / 64 	&\cellcolor{customcolor1} \textbf{5.89$_\text{+1.79}$}	&\cellcolor{customcolor1} 59.8&	\cellcolor{customcolor1} 66.0&	\cellcolor{customcolor1} 76.1&	\cellcolor{customcolor1} 68.9	&\cellcolor{customcolor1} 36.7	&\cellcolor{customcolor1} 63.6	&\cellcolor{customcolor1} 39.2&	\cellcolor{customcolor1} \textbf{58.6$_\text{-1.6}$} \\ 
        &  & \cellcolor{customcolor1} 24 / 64 	&\cellcolor{customcolor1} \textbf{14.95$_\text{+10.85}$}&	\cellcolor{customcolor1} 43.4	&\cellcolor{customcolor1} 65.8	&\cellcolor{customcolor1} 74.8	&\cellcolor{customcolor1} 67.1&	\cellcolor{customcolor1} 36.6	&\cellcolor{customcolor1} 62.9&	\cellcolor{customcolor1} 38.0	&\cellcolor{customcolor1} \textbf{55.5$_\text{-4.7}$} \\ 
    \midrule
        \multirow{4}{*}{\textbf{Zamba2-2.7B}} & Dense & 0 / 54 & 4.01&	69.7	&77.0	&79.8	&77.5&	48.5	&72.1&	45.8&	67.2  \\ 
        \cdashline{2-12}
        & \multirow{3}{*}{SSM Pruning} & \cellcolor{customcolor1} 16 / 54 	& \cellcolor{customcolor1} \underline{\textbf{4.14$_\text{+0.13}$}}	&\cellcolor{customcolor1} 69.2	&\cellcolor{customcolor1} 75.8&	\cellcolor{customcolor1} 79.2&	\cellcolor{customcolor1} 75.8&	\cellcolor{customcolor1} 46.5&	\cellcolor{customcolor1} 72.2	&\cellcolor{customcolor1} 45.8&	\cellcolor{customcolor1} \textbf{66.4$_\text{-0.8}$}\\ 
        & & \cellcolor{customcolor1} 20 / 54 	& \cellcolor{customcolor1} \underline{\textbf{5.07$_\text{+1.06}$}}	& \cellcolor{customcolor1} 64.2	&\cellcolor{customcolor1} 75.8&	\cellcolor{customcolor1} 79.3&	\cellcolor{customcolor1} 75.5&	\cellcolor{customcolor1} 46.2&	\cellcolor{customcolor1} 73.2&	\cellcolor{customcolor1} 46.0	&\cellcolor{customcolor1} \underline{\textbf{65.7$_\text{-1.5}$}} \\ 
        &  & \cellcolor{customcolor1} 24 / 54 &	\cellcolor{customcolor1} \underline{\textbf{5.46$_\text{+1.45}$}} &	\cellcolor{customcolor1} 62.3 &	\cellcolor{customcolor1} 74.7 &	\cellcolor{customcolor1} 79.0	 &\cellcolor{customcolor1} 75.4 &	\cellcolor{customcolor1} 44.3 &	\cellcolor{customcolor1} 70.9	 &\cellcolor{customcolor1} 46.4	 &\cellcolor{customcolor1} \underline{\textbf{64.7$_\text{-2.5}$}}  \\ 
    \bottomrule
    \end{tabular}
\caption{
Detailed results of \proj with \emph{training-free} SSM pruning. The remaining tasks represent their respective accuracy. Here, we do not consider the pruning ratio, as the number of SSM's parameter weights is small. Its benefit is the reduction of computational overhead. \underline{Underlined} numbers indicate the smallest gap with Dense under the same level of pruning.
}   
\label{tab:ssm_pruning}
\end{table*}

\subsection{Models} 
Our experiments employed the following pre-trained Mamba and hybrid models:  
\textbf{Mamba-2.8b}~\cite{mamba1}, consists of 64 S6 blocks\footnote{https://huggingface.co/state-spaces/mamba-2.8b}.  
\textbf{Mamba2-2.7b}~\cite{mamba2}, consists of 64 SSD blocks \footnote{https://huggingface.co/state-spaces/mamba2-2.7b}. Both Mamba models were trained on 300B tokens on the Pile dataset \cite{gao2020pile800gbdatasetdiverse}. 
For our choice of a hybrid model, we explored \textbf{Zamba2-2.7B}~\cite{glorioso2024zambacompact7bssm}\footnote{https://huggingface.co/Zyphra/Zamba2-2.7B}. It has 54 layers, including 45 single Mamba-2 Blocks and 9 hybrid layers composed of both Mamba-2 Blocks and Transformer Blocks. Zamba-2 was trained on 3T tokens from open web datasets, including Zyda \cite{tokpanov2024zyda}, and subsequently annealed with 100B additional tokens. 
The aforementioned models are all of the same size and can be compared directly. For Mamba models of different sizes, we also explored \textbf{Falcon-Mamba-7B}~\cite{zuo2024falcon}\footnote{https://huggingface.co/tiiuae/falcon-mamba-7b}, which is based on the Mamba-1 architecture and is the best-performing Mamba model at this scale in the literature, as well as \textbf{Hymba-1.5B-Base}~\cite{dong2024hymba}\footnote{https://huggingface.co/nvidia/Hymba-1.5B-Base}, which features a hybrid architecture incorporating both Mamba and Attention heads.

\subsection{Datasets}

Following the language modeling evaluation of Mamba \cite{mamba1, mamba2}, we utilize \emph{lm-eval-harness} \cite{eval-harness} to assess the zero-shot performance, which includes measuring perplexity on Lambada \cite{paperno-etal-2016-lambada}, and accuracy on the following downstream tasks: HellaSwag
\cite{zellers2019hellaswag}, Physical Interaction Question Answering (PIQA)
\cite{Bisk2020_piqa}, AI2 Reasoning Challenges (Arc-e, Arc-c)
\cite{Clark2018ThinkYH_arc}, Large-scale Winograd Schema Challenge (WinoGrande)
\cite{winogrande}, and the Open Book Question Answering \cite{Mihaylov2018CanAS_obqa} dataset.

Regarding the calibration dataset, we follow BlockPruner \cite{zhong2024blockprunerfinegrainedpruninglarge} in using the Alpaca dataset \footnote{https://github.com/tatsu-lab/stanford\_alpaca} as the calibration dataset and employ perplexity as the metric for calculating importance scores. All the hyperparameters used in our experiments are detailed in the Appendix.

\begin{table*}[!t]
    \setlength{\tabcolsep}{0.7pt}
    \centering
    \scriptsize
    \renewcommand\arraystretch{1.3}
    \begin{tabular}{lccccccccccc}
    \toprule
         \multirow{2}{*}{\textbf{Pruning Target}} & \textbf{Ratio} & \textbf{Additional} & \textbf{Lambada} & \multirow{2}{*}{\textbf{Lambada}} & \multirow{2}{*}{\textbf{HellaS}} & \multirow{2}{*}{\textbf{PIQA}}  & \multirow{2}{*}{\textbf{ARC-e}} & \multirow{2}{*}{\textbf{ARC-c}} & \multirow{2}{*}{\textbf{WinoG}}  & \multirow{2}{*}{\textbf{OBQA}} & \multirow{2}{*}{\textbf{Avg.}} \\ 
         & \textbf{(\textcolor{red}{Block}, \textcolor{red}{Width})} & \textbf{Pruned \textcolor{red}{SSMs}} & \textbf{PPL ($\downarrow$)} & & & & &  & & & \\ 
    \midrule
         / & 0\% & 0 / 54 & 4.01&	69.7	&77.0	&79.8	&77.5&	48.5	&72.1&	45.8&	67.2  \\ 
    \midrule
        Mamba Block \& Transformer Block & 10.40\% & 0 / 54 &	9.18$_\text{+5.17}$	&53.5	&67.3	&76.3	&63.5	&37.8	&64.3	&40.6	&57.6$_\text{-9.6}$ \\ 
        Mamba Block \& MLP \& MHA & 10.33\% & 0 / 54&	\textbf{5.01$_\text{+1.00}$}	&\textbf{65.6}	&73.6	&78.5	&75.3	&43.8	&69.3	&45.2	&64.5$_\text{-2.7}$ \\ 
        \cdashline{1-12}
        \cellcolor{customcolor1}Mamba Block \& MLP \& MHA + MLP Channel & \cellcolor{customcolor1}10.27\%  & \cellcolor{customcolor1} 0 / 54 &	\cellcolor{customcolor1} 5.45$_\text{+1.44}$&	\cellcolor{customcolor1} 63.4	&\cellcolor{customcolor1} \textbf{74.9}	&\cellcolor{customcolor1} \textbf{80.1}&	\cellcolor{customcolor1} \textbf{79.0}	&\cellcolor{customcolor1} \textbf{49.7}	&\cellcolor{customcolor1} \textbf{70.9}&	\cellcolor{customcolor1} 46.0	&\cellcolor{customcolor1} \textbf{66.3$_\text{-0.9}$} \\ 
        \cellcolor{customcolor1}Mamba Block \& MLP \& MHA + MLP Channel + SSM & \cellcolor{customcolor1}10.27\%  & \cellcolor{customcolor1} \textbf{18 / 54} &	\cellcolor{customcolor1} 5.18$_\text{+.1.17}$ &	\cellcolor{customcolor1} 63.4	&\cellcolor{customcolor1} 73.9	& \cellcolor{customcolor1} 80.0 &	\cellcolor{customcolor1} \textbf{79.0}	& \cellcolor{customcolor1} 48.7	& \cellcolor{customcolor1} 69.5 &	\cellcolor{customcolor1} \textbf{46.6}	& \cellcolor{customcolor1} 65.9$_\text{-1.3}$ \\ 

    \midrule
        Mamba Block \& Transformer Block & 15.89\%  & 0 / 54 &	10.38$_\text{+.6.37}$&	51.4	&65.6	&74.0	& 61.7	&37.7&	63.5&	39.6&	56.2$_\text{-11.0}$  \\ 
        Mamba Block \& MLP \& MHA & 15.54\%  & 0 / 54 &	10.64$_\text{+.6.63}$	&49.3&	69.2&	76.9	&66.1&	38.1	&66.0	&41.8	&58.2$_\text{-9.0}$ \\ 
        \cdashline{1-12}
        \cellcolor{customcolor1}Mamba Block \& MLP \& MHA + MLP Channel & \cellcolor{customcolor1}15.48\% & \cellcolor{customcolor1} 0 / 54 &	\cellcolor{customcolor1} \textbf{7.39$_\text{+.3.38}$}&	\cellcolor{customcolor1} \textbf{57.6}	&\cellcolor{customcolor1} \textbf{70.0}&	\cellcolor{customcolor1} \textbf{78.5}&	\cellcolor{customcolor1} \textbf{74.5}&	\cellcolor{customcolor1} \textbf{43.9}&	\cellcolor{customcolor1} 67.5&	\cellcolor{customcolor1} \textbf{43.8}&	\cellcolor{customcolor1} \textbf{62.3$_\text{-4.9}$} \\ 
        \cellcolor{customcolor1}Mamba Block \& MLP \& MHA + MLP Channel + SSM & \cellcolor{customcolor1}15.48\%  & \cellcolor{customcolor1} \textbf{18 / 54} &	\cellcolor{customcolor1} 7.43$_\text{+.3.42}$ &	\cellcolor{customcolor1} 56.5 &	\cellcolor{customcolor1} 68.9	 &\cellcolor{customcolor1} 77.9 &	\cellcolor{customcolor1} 73.4	 &\cellcolor{customcolor1} 41.8	 &\cellcolor{customcolor1} \textbf{67.7}	 &\cellcolor{customcolor1} 42.8 &	\cellcolor{customcolor1} 61.3$_\text{-5.9}$ \\ 
    \bottomrule
    \end{tabular}
\caption{
Results of Zamba2-2.7B were achieved by pruning its Mamba-2 and Transformers blocks at multiple granularities, including entire Mamba-2 block, MHA block, MLP block, MLP channel, and SSM module. The remaining tasks represent their respective accuracies. ``\&'' indicates that the pruning targets are considered together in the same pruning step, while ``+'' signifies the distinction between pruning stages, with pruning occurring sequentially. \textbf{Bold} numbers indicate the best performance under the same level of pruning (excluding Dense).
}
\label{tab:zamba_multi_pruning}
\end{table*}
\vspace{-3pt}
\subsection{Results}
\vspace{-2pt}
\subsubsection{Pruning Target: Mamba Block}
\vspace{-1pt}
This section explores the impact of pruning Mamba blocks on model performance. Figure \ref{fig:mamba_block_pruning} and Table \ref{tab:mamba_block_pruning} present the results of applying \proj to Mamba-2.8B, Mamba2-2.7B, and Zamba2-2.7B models with a focus on removing redundant entire Mamba blocks. The model that utilizes the first version of Mamba blocks (S6) appears to tolerate a higher number of removed blocks without significantly affecting its performance. Specifically, the Mamba-2.8B model demonstrates robustness, with its perplexity (PPL) increasing from 4.23 to 7.51 and average accuracy dropping from 59.9 to 53.8 when the pruning ratio reaches 20.86\%. In contrast, the Mamba2-2.7B and Zamba2-2.7B models exhibit more significant performance degradation, although they performed better before pruning (Dense). The poorer pruning performance of Zamba2-2.7B may be attributed to the pruning of Mamba blocks disrupting a certain balance within the hybrid layers. Overall, the effects of Mamba block pruning vary across different models, depending on the model architecture and the characteristics of the pre-training stage. In this round, Mamba-1 comes out on top.

\subsubsection{Pruning Target: SSM Module}
\label{sec:pruning_target_ssm_module}
\vspace{-2pt}
In this section, we delve into assessing the impact of pruning only the SSM modules within Mamba blocks on the performance of various models, as illustrated in Table \ref{tab:ssm_pruning} and Figure \ref{fig:ssm_pruning}.
When using the same target in Mamba-2.8B, we observe that further pruning SSMs results in a noticeable increase in perplexity, soaring to 22.55 and decreasing average accuracy to 44.1. This result indicates a significant sensitivity to SSM pruning for Mamba-1, where performance degradation is pronounced even at moderate pruning levels.
Conversely, Mamba2-2.7B and Zamba2-2.7B exhibit remarkable resilience to SSM pruning. Even with 24 SSMs pruned, the model maintains a relatively stable performance. This robustness suggests that Mamba-2 blocks can tolerate higher SSM module pruning, potentially due to Mamba-2's optimizations or different training strategies with Mamba-1.
The Zamba2-2.7B model, with the hybrid architecture, outperforms both Mamba-1 and Mamba-2. Pruning 12 out of its 54 SSMs results in a negligible PPL increase from 4.01 to 4.02, while the average accuracy slightly decreases from 67.2\% to 67.0\%. The hybrid nature of Zamba2-2.7B may contribute to its ability to maintain performance despite SSM pruning.
Overall, these findings underscore the importance of model architecture and training strategies in determining the impact of SSM pruning. They offer valuable insights for optimizing model efficiency without compromising performance. In this round, the model with Mamba-2 blocks comes out on top.


\subsubsection{Pruning Target: Finer-grained removal of Mamba and Transformer blocks, and their subcomponents}
\vspace{-2pt}

Table \ref{tab:zamba_multi_pruning} presents the results of pruning various components of the Zamba2-2.7B model, including combinations of Mamba-2 blocks, entire Transformer blocks, and their subcomponents, i.e., MHA blocks, MLP blocks, MLP channels, and SSM modules. 
We design four search spaces to study the effectiveness of different granularities and their combinations. ``\&'' indicates that the pruning targets are considered together in the same pruning step, while ``+'' signifies the distinction between pruning stages, with pruning occurring sequentially:
\vspace{-3pt}
\paragraph{Mamba Block \& Transformer Block Pruning} This experiment involves pruning the entire Mamba-2 blocks and Transformer blocks.
\vspace{-3pt}
\paragraph{Mamba Block \& MLP \& MHA Pruning} This experiment decomposes the transformer block into sub-blocks, pruning Mamba-2 blocks as well as MHA and MLP.
\vspace{-3pt}
\paragraph{Mamba Block \& MLP \& MHA + MLP Channel Pruning} This experiment prunes the Mamba-2 blocks, MHA, and MLP at the first stage and further prunes the MLP channels at the next stage.
\vspace{-3pt}
\paragraph{Mamba Block \& MLP \& MHA + MLP Channel Pruning + SSM} Add additional SSM pruning following the previous solution.

The results indicate that pruning Mamba blocks and Transformer blocks alone leads to significant performance degradation. However, \textbf{more granular pruning strategies show a more favorable trade-off between pruning ratio and performance}. Specifically, pruning Mamba blocks, MLP, MHA (single stage), and MLP channels subsequently performs the best. Inspired by the SSM pruning of Mamba-2 in Section \ref{sec:pruning_target_ssm_module}, we further add SSM pruning to the third strategy, and the results show that removing around 18 SSMs can maintain accuracy performance while reducing computational overhead. An interesting finding is that pruning SSMs can even lower PPL; for instance, at a 10\% pruning ratio, PPL decreases from 5.45 to 5.18, suggesting that some SSM modules are redundant after the second pruning stage. Overall, these findings indicate that multi-granularity pruning methods, particularly those including MLP channels and SSM modules, can effectively reduce the complexity of hybrid Mamba models while maintaining a higher level of performance.

\begin{table*}[!t]
    \setlength{\tabcolsep}{2.5pt}
    \centering
    \scriptsize
    \renewcommand\arraystretch{1.2}
    \begin{tabular}{llccccccc}
    \toprule
        \multirow{2}{*}{\textbf{Model}} & \multirow{2}{*}{\textbf{Method}} & \textbf{Num. of Pruned}  & \multirow{2}{*}{\textbf{HellaS}} & \multirow{2}{*}{\textbf{PIQA}}  & \multirow{2}{*}{\textbf{ARC-e}} & \multirow{2}{*}{\textbf{ARC-c}} & \multirow{2}{*}{\textbf{WinoG}}   & \multirow{2}{*}{\textbf{Average}} \\ 
         &  & \textbf{\textcolor{red}{Hymba Blocks}}  \\ 
    \midrule
        \multirow{4}{*}{\textbf{Hymba-1.5B-Base}} & Dense & 0 / 32  & 53.5 & 77.1 & 76.6 & 45.4 & 66.1 & 63.8 \\ 
        \cdashline{2-9}
        & \multirow{3}{*}{Hymba Block Pruning} & 6 / 32 & 50.5 & 75.8 & 76.0 & 44.9 & 64.1 & 62.3  \\ 
        &  & 7 / 32 & 49.9 & 74.9 & 74.8 & 43.9 & 64.9 & 61.7 \\ 
        &  & 8 / 32 &49.2 & 74.3 & 74.2 & 43.2 & 61.5& 60.5 \\ 
    \bottomrule
    \end{tabular}
\caption{
Results of \proj with \emph{training-free} Hymba block pruning for Hymba-1.5B-Base \cite{dong2024hymba}. Five commonsense reasoning tasks are used for evaluation: HellaSwag, PIQA, ARC-e, ARC-c, and WinoGrande.}
\label{tab:hymba_results}
\end{table*}

\subsubsection{Pruning Mamba Models of Other Sizes}

\paragraph{Hymba}
Table \ref{tab:hymba_results} shows the results of \proj with training-free Hymba Block pruning for Hymba-1.5B-Base. The dense configuration achieves an average accuracy of 63.8, which decreases as more blocks are pruned, dropping to 60.5 when 8 blocks are pruned, indicating a general decline in performance across benchmarks. Further analysis of inference acceleration and recovery tuning experiments for Hymba-1.5B-Base will be discussed in the subsequent sections.

\paragraph{Falcon-Mamba}

\begin{table*}[!t]
    \setlength{\tabcolsep}{2.5pt}
    \centering
    \scriptsize
    \renewcommand\arraystretch{1.2}
    \begin{tabular}{llcccccccccc}
    \toprule
        \multirow{2}{*}{\textbf{Model}} & \multirow{2}{*}{\textbf{Method}} & \textbf{Num. of Pruned} & \textbf{Lambada} & \multirow{2}{*}{\textbf{Lambada}} & \multirow{2}{*}{\textbf{HellaS}} & \multirow{2}{*}{\textbf{PIQA}}  & \multirow{2}{*}{\textbf{ARC-e}} & \multirow{2}{*}{\textbf{ARC-c}} & \multirow{2}{*}{\textbf{WinoG}}  & \multirow{2}{*}{\textbf{OBQA}} & \multirow{2}{*}{\textbf{Average}} \\ 
         &  & \textbf{\textcolor{red}{Mamba Blocks} / \textcolor{red}{SSMs}}& \textbf{PPL ($\downarrow$)}  \\ 
    \midrule
        \multirow{9}{*}{\textbf{Falcon-Mamba-7B}} & Dense & 0 / 64 & 3.15&	74.3	&80.3	&82.0	&84.4&	58.9&	75.1&	49.0&	72.0  \\ 
        \cdashline{2-12}
        & \multirow{4}{*}{Mamba Block Pruning} & 5 / 64 &  4.01&	69.2&	78.6&	81.9	&82.2&	54.6&	72.5&	47.6	&\textbf{69.5} \\ 
        &  & 10 / 64  & 4.97&	65.1&	75.0	&79.5	&79.7&	51.5&	70.2&	43.8	&\textbf{66.4}\\ 
        &  & 15 / 64 & 5.63&	62.4&	71.2	&77.8	&76.1	&49.1&	70.2&	41.8	&\textbf{64.1} \\ 
        &  & 20 / 64 & 39.31&	31.5&	65.9&	74.3	&72.2&	42.3	&65.2&	38.4	&\textbf{55.7} \\ 
        \cdashline{2-12}
        & \multirow{4}{*}{SSM Pruning} &  5 / 64 & \textbf{3.47}	&71.6	&77.3	&81.2&	77.8	&49.2&	73.2&	47.2&	68.2   \\ 
        &  & 10 / 64 & \textbf{4.24}&67.2	&73.6&	79.8	&75.3&	48.3&	70.2&	43.0	&65.4 \\ 
        &  & 15 / 64 & \textbf{5.37}	&63.3&	69.6	&78.2	&72.4	&43.4	&68.8	&41.8	&62.5  \\ 
        &  & 20 / 64 & \textbf{14.14}&	46.3	&63.4&	74.9&	60.7	&36.7	&65.7	&37.8&	55.1\\  
    \bottomrule
    \end{tabular}
\caption{
Results of \proj with \emph{training-free} Mamba block pruning and SSM pruning for Falcon-Mamba-7B.
}   
\label{tab:falcon_mamba_7b_results}
\end{table*}

While the previous sections focused on exploring the pruning of Mamba models with sizes around 2.7B or 2.8B, we also investigated the impact of \proj on a larger-scale Mamba model, specifically Falcon-Mamba-7B (Table \ref{tab:falcon_mamba_7b_results}).
Pruning SSM modules in the Falcon-Mamba-7B model shows better tolerance in terms of perplexity, suggesting that SSM pruning is more effective in maintaining lower perplexity. Regarding average accuracy, pruning entire Mamba blocks is more beneficial. 

Additionally, it is important to note that pruning entire Mamba blocks yields more significant computational benefits than SSM pruning, suggesting that while SSM pruning is advantageous for maintaining perplexity, pruning Mamba blocks offers a better trade-off between computational efficiency and accuracy. 
The choice of pruning strategy should be guided by the specific performance metric of interest and the desired balance between computational efficiency and model accuracy.

None of the above results have undergone fine-tuning to improve the performance of the pruned models. As in many other works, the drop in the accuracy performance of pruned models can be recovered by fine-tuning, which will be incorporated in Section \ref{sec:recovery_tuning}.

\subsection{Inference Acceleration}

\begin{table}[!t]
\setlength{\abovecaptionskip}{0.2pt}
\setlength{\tabcolsep}{3.5pt}
\caption{Inference benchmark results for Mamba-2.8B. The batch size is 1. Number of batches is 10. The prompt length is 512. Number of new tokens is 16. }
\label{tab:inference_speedup_mamba}
\scriptsize
\centering
\renewcommand\arraystretch{1.3}
\begin{tabular}{llccc}
\toprule
 \multirow{2}{*}{\textbf{Model}} &\multirow{2}{*}{\textbf{Method}} & \textbf{Num. of Pruned} & \multicolumn{2}{c}{\textbf{Inference Speedup}}   \\
\cdashline{4-5}
 & & \textbf{\textcolor{red}{Mamba Blocks}} & \textbf{Prefill} & \textbf{Decode}  \\ 
\midrule
 \multirow{3}{*}{\textbf{Mamba-2.8B}} & Dense & 0 / 64   & 1.00$\times$ & 1.00$\times$ \\ 
 \cdashline{2-5}
& \multirow{2}{*}{\proj}& 7 / 64      & \textbf{1.12$\times$} & \textbf{1.13$\times$} \\
& & 14 / 64      & \textbf{1.31$\times$} & \textbf{1.29$\times$} \\
\bottomrule
\end{tabular}
\end{table}

\begin{table}[!t]
\setlength{\abovecaptionskip}{1pt}
\setlength{\tabcolsep}{4pt}
\caption{Inference benchmark results for Mamba2-2.7B, with test-related hyperparameters consistent with Table \ref{tab:inference_speedup_mamba}.
}
\label{tab:inference_speedup_mamba2}
\scriptsize
\centering
\renewcommand\arraystretch{1.3}
\begin{tabular}{llccc}
\toprule
\multirow{2}{*}{\textbf{Model}} & \multirow{2}{*}{\textbf{Method}} & \textbf{Num. of} & \multicolumn{2}{c}{\textbf{Inference Speedup}}   \\
\cdashline{4-5}
&  & \textbf{Pruned \textcolor{red}{SSMs}} & \textbf{Prefill} & \textbf{Decode}  \\ 
\midrule
 \multirow{4}{*}{\textbf{Mamba2-2.7B}} & Dense & 0 / 64   & 1.00$\times$ & 1.00$\times$ \\ 
 \cdashline{2-5}
& \multirow{3}{*}{\proj} & 16 / 64    & \textbf{1.13$\times$} & \textbf{1.11$\times$} \\
&  & 20 / 64    & \textbf{1.16$\times$} & \textbf{1.14$\times$} \\
&  & 24 / 64    & \textbf{1.20$\times$} & \textbf{1.18$\times$} \\
\bottomrule
\end{tabular}
\end{table}

\begin{table}[!t]
\setlength{\tabcolsep}{1pt}
\caption{Inference benchmark results for Zamba2-2.7B, with test-related hyperparameters consistent with Table \ref{tab:inference_speedup_mamba}. The calculation of \emph{Ratio} includes block pruning (Mamba Block, MHA, and MLP) and width pruning (MLP Channel). Refer to Table \ref{tab:zamba_multi_pruning} for more information.}
\label{tab:inference_speedup_zamba2}
\scriptsize
\centering
\renewcommand\arraystretch{1.3}
\begin{tabular}{llccccc}
\toprule
 \multirow{2}{*}{\textbf{Model}} &\multirow{2}{*}{\textbf{Method}} & \textbf{Ratio} & \textbf{Additional} & \multicolumn{2}{c}{\textbf{Inference Speedup}}   \\
\cdashline{5-6}
 & & \textbf{(\textcolor{red}{Block}, \textcolor{red}{Width})} & \textbf{Pruned \textcolor{red}{SSMs}} & \textbf{Prefill} & \textbf{Decode}  \\ 
\midrule
 \multirow{2}{*}{\textbf{Zamba2}} & Dense & 0\% & 0 / 54   & 1.00$\times$ & 1.00$\times$ \\ 
 \cdashline{2-6}
\multirow{2}{*}{\textbf{-2.7B}} & \multirow{2}{*}{\proj} & 15.48\% & 0 / 54    & \textbf{1.16$\times$} & \textbf{1.34$\times$} \\
& & 15.48\% & 18 / 64    & \textbf{1.25$\times$} & \textbf{1.39$\times$} \\
\bottomrule
\end{tabular}
\end{table}

\vspace{-2pt}
Through the above analysis, we have gained a good understanding and insight into the impact of \proj's structured pruning on model accuracy and perplexity performance.
In addition, through structured pruning, \proj achieves an additional speedup to these already highly efficient models. Next, we discuss the impact of inference acceleration. All the following tests were conducted on a single Tesla V100 32GB GPU.
\vspace{-2pt}
\paragraph{Mamba-1} When removing entire Mamba blocks, as shown in Table \ref{tab:inference_speedup_mamba}, \proj speeds up the decoding stage up to 1.29x when removing 14 blocks, and 1.13x when removing only 7 blocks, which highlights the potential of \proj to optimize computational efficiency in Mamba models. The user's decision on how aggressively to prune will impact the average accuracy or the perplexity as observed in Table \ref{tab:mamba_block_pruning}.
\vspace{-2pt}
\paragraph{Mamba-2} As detailed in Table \ref{tab:inference_speedup_mamba2}, removing 24 SSM modules (~44\% of the total number of modules) results in up to a 1.20x speedup in the prefill stage and a 1.18x speedup in the decoding stage during of inference. 
A more conservative pruning ratio achieves 1.11x speedup when removing 16 SSM modules. Based on previous observations, the impact on performance metrics is minimal (0.4\% for accuracy and 0.16 for PPL). These results underscore the effectiveness of SSM pruning in enhancing computational efficiency while barely affecting model performance, making it a viable strategy for optimizing Mamba models.

\paragraph{Zamba-2}
As detailed in Table \ref{tab:inference_speedup_zamba2}, we observe significant acceleration on inference after multiple granularities pruning of Zamba-2. Specifically, pruning Mamba blocks, MLP, and MHA blocks along with MLP channels results in a 1.34x speedup in the decoding stage. When SSM pruning is included, the speedup increases to 1.39x, indicating that a comprehensive pruning strategy that includes multiple components can significantly enhance inference speed while maximizing the preservation of model performance.

\paragraph{Hymba} As shown in Table \ref{tab:inference_speedup_hymba}, the hymba block pruning of Hymba-1.5B-Base demonstrates notable improvements in inference speed. By removing 7 out of 64 Hymba blocks, \proj achieves a 1.15x speedup in the prefill stage and a 1.24x speedup in the decoding stage,  suggesting that significant computational efficiency gains can be realized even with a relatively modest pruning ratio. The results highlight the potential of \proj to optimize the performance of Hymba models, making them more efficient for real-time applications without substantial sacrifices in model accuracy.

\begin{table}[!t]
\setlength{\abovecaptionskip}{1pt}
\setlength{\tabcolsep}{2pt}
\caption{Inference benchmark results for Hymba-1.5B-Base, where the test-related hyperparameters consistent with Table \ref{tab:inference_speedup_mamba}, except that number of new tokens is 256.
}
\label{tab:inference_speedup_hymba}
\scriptsize
\centering
\renewcommand\arraystretch{1.3}
\begin{tabular}{llccc}
\toprule
  \multirow{2}{*}{\textbf{Model}} & \multirow{2}{*}{\textbf{Method}} & \textbf{Num. of Pruned} & \multicolumn{2}{c}{\textbf{Inference Speedup}}   \\
\cdashline{4-5}
 &  & \textbf{\textcolor{red}{Hymba Blocks}} & \textbf{Prefill} & \textbf{Decode}  \\ 
\midrule
\multirow{2}{*}{\textbf{Hymba-1.5B-Base}}& Dense & 0 / 64   & 1.00$\times$ & 1.00$\times$ \\ 
 \cdashline{2-5}
&\multirow{1}{*}{\proj} & 7 / 64    & \textbf{1.15$\times$} & \textbf{1.24$\times$} \\
\bottomrule
\end{tabular}
\end{table}

\subsection{Recovery Tuning of the Pruned Model}
\label{sec:recovery_tuning}

\begin{table}[!t]
    \setlength{\tabcolsep}{1.5pt}
    \centering
    \scriptsize
    \renewcommand\arraystretch{1.3}
    \begin{tabular}{llccc}
    \toprule
         \multirow{2}{*}{\textbf{Model}} &\multirow{2}{*}{\textbf{Method}} & \textbf{Num. of} & \textbf{Lambada} & \textbf{Average} \\ 
        & &  \textbf{Pruned \textcolor{red}{SSMs}} & \textbf{PPL ($\downarrow$)} & \textbf{Accuracy} \\ 
    \midrule
        \multirow{3}{*}{\textbf{Mamba2-2.7B}} &  Dense &  0 / 64 & 4.10&	60.2  \\ 
        \cdashline{2-5}
        & \proj &  20 / 64 &	5.89 & 58.6 \\ 
        & \cellcolor{customcolor2}\proj \textbf{w/ tune}  & \cellcolor{customcolor2}20 / 64 & \cellcolor{customcolor2}\textbf{4.44$_\text{-1.45}$}& \cellcolor{customcolor2}\textbf{59.6$_\text{+1.0}$} \\ 
    \bottomrule
    \end{tabular}
\caption{Results of the compressed Mamba2-2.7B model with recovery tuning (post-training).
}
\label{tab:recovery_tuning_mamba2}
\end{table}

\begin{table*}[!t]
    \setlength{\tabcolsep}{4.5pt}
    \centering
    \scriptsize
    \renewcommand\arraystretch{1.3}
    \begin{tabular}{llcccc}
    \toprule
         \multirow{2}{*}{\textbf{Model}} &\multirow{2}{*}{\textbf{Method}} & \textbf{Ratio} & \textbf{Additional} & \textbf{Lambada} & \textbf{Average} \\ 
         & & \textbf{(\textcolor{red}{Block}, \textcolor{red}{Width})} & \textbf{Pruned \textcolor{red}{SSMs}} &\textbf{PPL ($\downarrow$)} & \textbf{Accuracy} \\ 
    \midrule
        \multirow{5}{*}{\textbf{Zamba2-2.7B}} & Dense & - & 0 / 54 & 4.01&	67.2  \\ 
        \cdashline{2-6}
        & \proj & 10.27\% & 18 / 54 &	5.18& 65.9 \\ 
        &\cellcolor{customcolor2}\proj \textbf{w/ tune} & \cellcolor{customcolor2}10.27\% & \cellcolor{customcolor2}18 / 54 & \cellcolor{customcolor2}\textbf{4.58$_\text{-0.60}$}& \cellcolor{customcolor2}\textbf{67.0$_\text{+1.1}$} \\
        \cdashline{2-6}
        &\proj & 15.48\%  & 18 / 54 &	7.43 &	61.3 \\ 
        &\cellcolor{customcolor2}\proj \textbf{w/ tune} & \cellcolor{customcolor2}15.48\%  & \cellcolor{customcolor2}18 / 54 &	\cellcolor{customcolor2}\textbf{5.88$_\text{-1.55}$} &	\cellcolor{customcolor2}\textbf{64.4$_\text{+3.1}$} \\ 
    \bottomrule
    \end{tabular}
\caption{Results of the compressed Mamba2-2.7B and Zamba2-2.7B models with recovery tuning.
}
\label{tab:recovery_tuning_zamba2}
\end{table*}

\begin{table}[!t]
    \setlength{\tabcolsep}{2pt}
    \centering
    \scriptsize
    \renewcommand\arraystretch{1.3}
    \begin{tabular}{llcc}
    \toprule
         \multirow{2}{*}{\textbf{Model}} &\multirow{2}{*}{\textbf{Method}} & \textbf{Num. of Pruned} & \textbf{Average} \\ 
        & &  \textbf{\textcolor{red}{Hymba Blocks}} &  \textbf{Accuracy} \\ 
    \midrule
        \multirow{3}{*}{\textbf{Hymba-1.5B-Base}} &  Dense &  0 / 32 &	63.8  \\ 
        \cdashline{2-4}
        & \proj &  7 / 32  & 61.7 \\ 
        & \cellcolor{customcolor2}\proj \textbf{w/ tune}  & \cellcolor{customcolor2}7 / 32 & \cellcolor{customcolor2}\textbf{63.7$_\text{+2.0}$} \\ 
    \bottomrule
    \end{tabular}
\caption{Results of the compressed Hymba-1.5B-Base model with recovery tuning. \emph{Average Accuracy} is calculated over HellaSwag, PIQA, ARC-e, ARC-c, and WinoGrande tasks (Table \ref{tab:hymba_results}).
}
\label{tab:recovery_tuning_hymba}
\end{table}

\vspace{-1pt}
Following most of the work \cite{ma2023llmpruner, zhong2024blockprunerfinegrainedpruninglarge}, we performed post-training on the \proj compressed model using the cleaned version of Alpaca. The results summarized in Tables \ref{tab:recovery_tuning_mamba2}, \ref{tab:recovery_tuning_zamba2}, and \ref{tab:recovery_tuning_hymba} demonstrate substantial performance gains after just two epochs of recovery tuning (see Appendix for more hyperparameters).
For instance, the \proj model obtained by removing Mamba Blocks \& MLPs \& MHAs + MLP Channels + SSM in Zamba-2 (Table \ref{tab:zamba_multi_pruning}), initially exhibits a perplexity of 5.18 and an average accuracy of 65.9 when 18 out of 54 SSMs are pruned. However, after recovery tuning, it achieves a significantly reduced PPL of 4.58 and an improved average accuracy of 67.0, which is almost on par with the Dense model. 
Similarly, as shown in Table \ref{tab:recovery_tuning_hymba}, the recovery tuning of the Hymba-1.5B-Base model also yields significant improvements. Initially, the pruned model with 7 out of 32 Hymba blocks removed shows an average accuracy of 61.7. After recovery tuning, the average accuracy increases to 63.7, which is nearly equivalent to the dense model's accuracy of 63.8. 
These results indicate that the recovery fine-tuning phase effectively enhances the performance of the pruned model, bringing it closer to the original dense model's performance while maintaining computational efficiency. 
In summary, recovery tuning is crucial to optimize pruned models, making them more viable for practical applications.
\vspace{-0.2cm}
\subsection{Insights on the Compression Sensitivity of the Variants of Mamba}
\vspace{-1pt}

\begin{figure}
  \centering
  \begin{minipage}{0.23\textwidth}
        \centering
        \includegraphics[width=\textwidth]{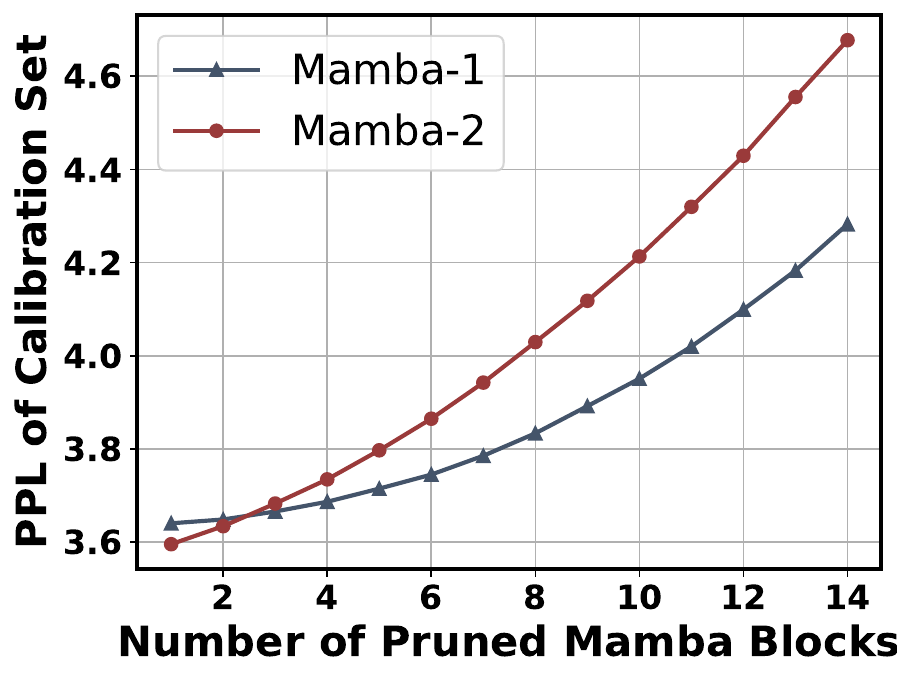}
    \end{minipage}
    \begin{minipage}{0.23\textwidth}
        \centering
        \includegraphics[width=\textwidth]{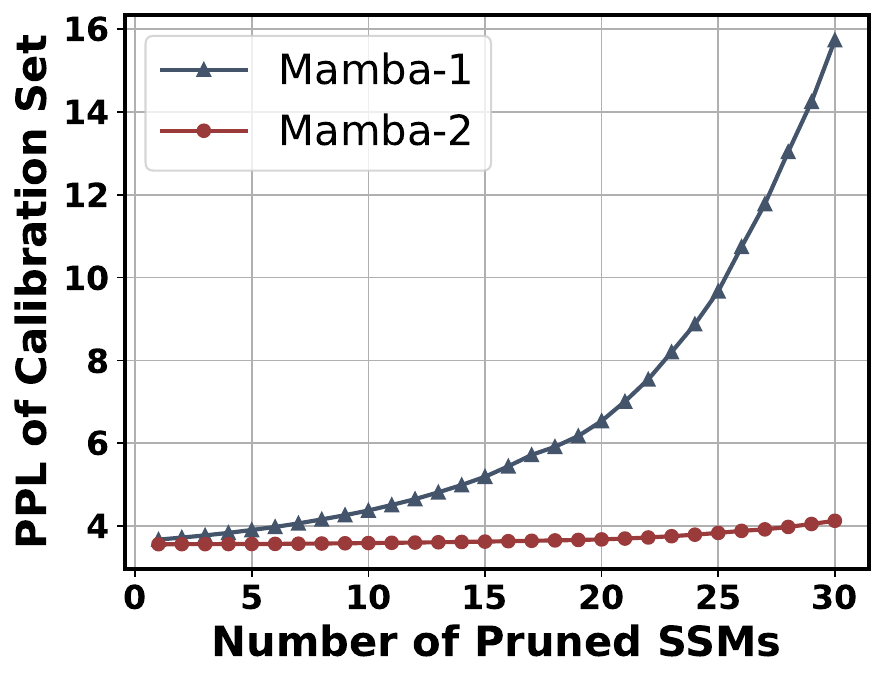}
    \end{minipage}
  \caption{Close examination of the impact of removing Mamba blocks or SSMs from the two versions of Mamba models reveals distinct differences in their tolerance levels. Mamba-1 exhibits a higher tolerance for removing its blocks, while Mamba-2 exhibits greater tolerance for removing the SSM subcomponent.  
  }
\label{fig:mamba_sensitivity}
\end{figure}

A research question during our investigation considered, \emph{will the improvements in Mamba-2 make it more sensitive to removing its inner structures?} 

The proponents of Mamba modified the original architecture to restrict the expressivity in Mamba-2 and increase the training efficiency. As illustrated on the left side of Figure \ref{fig:mamba_sensitivity}, our experiments suggest that these changes make Mamba-2 models less robust to removing entire blocks than the previous version of the Mamba block. As soon as we remove blocks with the least importance, Mamba-1 exhibits a more robust behavior. However, Mamba-2 demonstrates a significantly higher tolerance to removing SSMs, maintaining a stable perplexity even as more SSMs are pruned, suggesting that while Mamba-2's architectural improvements have made it more sensitive to the removal of Mamba blocks, they have also enhanced its robustness to SSM pruning.

\vspace{-0.2cm}
\section{Conclusion}
\label{sec:conclusion}
\vspace{-0.2cm}
Selective structure state space models have become an efficient alternative to Transformer-based models. In this paper, we propose \proj and investigate structured pruning strategies to remove elements from Mamba and hybrid models and reduce model size, accelerating inference. The results demonstrate that selective structured state space architectures have several redundancies that can be removed without significantly affecting the model's performance.

\section*{Limitations} 

Despite their outstanding results, large sequence models are still under investigation to better understand their capabilities and limitations. \proj is, to the best of our knowledge, the first work to investigate the removal of structures in Mamba-based models, including hybrids with Transformer blocks. Our goal is to motivate the research community to better understand this class of models to identify opportunities for future improvements in the model architecture and applicable compression techniques. The results indicate that these models contain redundant elements that might be removed to improve their efficiency. However, future work must explore and attempt to better understand the trade-offs between efficiency and accuracy when removing these models' components. Even more research questions can be entertained when considering Transformer blocks and hybrid models, as in the case of Zamba. For instance, there is much to understand about the right mix of the SSM- and Transformer-based elements. 

\section*{Ethics Statement}

Due to the well-known flaws in modern sequence models, e.g., hallucinations, many guard rails must be in place when considering deploying them in production. Our research focuses on improving the efficiency of these models in existing downstream tasks and datasets. However, further experimentation and analysis are needed when considering deploying these compressed models in environments where their output might affect people's well-being. 

\section*{Acknowledgments}

We are grateful to Michael Beale from Intel Labs, who helped us set up the infrastructure for sharing our models during the review stage and the final release and guided us through open-sourcing our compressed models. We also thank the anonymous reviewers for their insightful suggestions, which helped us improve the paper.

\bibliography{main}

\clearpage
\appendix
\section*{Supplementary Material} 
\label{sec:appendix}

\section{Related Work}

Transformers \cite{Transformer_NIPS2017} and its variants are the primary building block of successful deep learning architectures, e.g., Llama \cite{touvron2023llama2} and GPT \cite{NEURIPS2020_1457c0d6_GPT}, that have revolutionized Natural Language Processing (NLP) \cite{devlin2019bert, eval-harness}, Computer Vision (CV) \cite{46840_image_transformer, radford2021clip, SETR-HLG_visiontrans}, and many other domains. Due to the Transformer's popularity, researchers have proposed variants to improve their computational and memory efficiency further and tackle issues like their quadratic complexity in sequence length during training \cite{correia-etal-2019-adaptively, Beltagy2020Longformer,  10.5555/3495724.3496083_funneltransformer, choromanski2021rethinking_Performer, 10.5555/3524938.3525416_linearattention, zheng2022linear}. 

A parallel research effort investigates alternatives to Transformers in the form of \emph{structured state space models} (SSMs) that can power the next generation of sequence models. The initial proposals of structured SSMs were linear time-invariant, e.g., LSSL \cite{10.5555/3540261.3540305_lssl}, S4 \cite{gu2022efficientlyS4}, H3 \cite{fu2023hungry}. Recent improvements to the state space model formulation have resulted in the proposal of time-varying selective SSMs, e.g., Mamba \cite{mamba1, mamba2}. 

To our knowledge, \proj is the first study on pruning selective structured state space models (Mamba) and their hybrids. On the other hand, many works have proposed pruning techniques for Transformer-based models \cite{hoeflerSparsity21}. Several of these works focus on \emph{unstructured} pruning \cite{sun2023wanda, xu2024besa, frantar-gptq}, which can achieve higher sparsity levels. However, it requires highly optimized runtimes to realize the benefits of sparsity. Sophisticated solutions have been proposed to fine-tune sparse models and recover any accuracy drop from the pruning stage \cite{munoz-etal-2024-shears}. Recently, \emph{training-free} approaches have been proposed for \emph{structured} pruning of Transformers. These approaches cannot achieve high sparsity levels as the \emph{unstructured} pruning approaches. However, they are very convenient because their compressed models do not require specialized runtimes and exhibit beneficial inference acceleration. In this line of research, LLMPruner \cite{ma2023llmpruner}, ShortGPT \cite{men2024shortgptlayerslargelanguage}, BlockPruner \cite{lagunas-etal-2021-block}, SliceGPT \cite{ashkboos2024slicegpt}, and MultiPruner \cite{muñoz2025multiprunerbalancedstructureremoval} have demonstrated efficient methods for Transformer pruning. BlockPruner improved over many previous approaches by proposing a global metric that can be used to determine the importance of a selected network structure. MultiPruner extended this approach to pruning the width dimension, as well. \proj builds on these works and the rest of the extensive literature on \emph{structured} block pruning to explore opportunities for removing redundancies in models with Mamba blocks. 

\section{Hyperparameters}

Table \ref{tab:hyperparameters} offers a detailed summary of the hyperparameters employed in our experiments, promoting both reproducibility and clarity.

\begin{table}[!t]
    \setlength{\tabcolsep}{2pt}
    \renewcommand\arraystretch{1.2}
    \centering
    \footnotesize 
    \begin{tabular}{lcc}
    \toprule
\textbf{Hyper-parameter} & \textbf{Value} \\
\midrule
\textit{Pruning Stage:} \\
\cdashline{1-2}
Calibration Dataset & tatsu-lab/alpaca  \\
Importance Metric & Perplexity (PPL)  \\
Number of Calibration Samples & 256  \\
MLP Channel Group Size (Zamba2) & 1024  \\
Steps of MLP Channel Pruning (Zamba2) & 20  \\
\bottomrule
\end{tabular}
\caption{Hyper-parameters used in the experiments. 
}
\label{tab:hyperparameters}
\end{table}

\end{document}